\documentclass[10pt,twocolumn,letterpaper]{article}

\usepackage[final]{wacv}      

\usepackage{graphicx}
\usepackage{amsmath}
\usepackage{amssymb}
\usepackage{booktabs}
\usepackage{xspace}
\newcommand{\ours}{PRN\xspace}
\usepackage[dvipsnames]{xcolor}
\usepackage{booktabs}
\usepackage{enumitem}
\usepackage{multirow}
\usepackage{tabularx}
\usepackage{hhline}
\usepackage[makeroom]{cancel}

\usepackage[pagebackref,breaklinks,colorlinks]{hyperref}
\usepackage{comment}

\usepackage[capitalize]{cleveref}
\crefname{section}{Sec.}{Secs.}
\Crefname{section}{Section}{Sections}
\Crefname{table}{Table}{Tables}
\crefname{table}{Tab.}{Tabs.}

\def\wacvPaperID{2275} 
\def\confName{WACV}
\def\confYear{2024}

\DeclareMathOperator{\threshold}{\tau}
\newcommand{\numpatches}{\ensuremath{k}}

\DeclareMathOperator{\loss}{\textbf{L}}
\DeclareMathOperator{\softmax}{softmax}
\DeclareMathOperator{\abs}{abs}
\DeclareMathOperator{\conv}{conv}
\DeclareMathOperator{\numclasses}{m}

\begin{document}

\title{PatchRefineNet: Improving Binary Segmentation by Incorporating Signals from Optimal Patch-wise Binarization}

\author{Savinay Nagendra $^{1}$  \quad Chaopeng Shen$^{2}$ \quad Daniel Kifer$^{1}$ \vspace{0.3em} \\
{\normalsize $^1$Department of Computer Science} \quad
{\normalsize $^2$Department of Civil and Environmental Engineering} \quad \\
{\normalsize The Pennsylvania State University}\\
{\normalsize University Park}\\
{\tt\small\centering $^1$sxn265@psu.edu \quad $^2$cshen@engr.psu.edu \quad $^1$dkifer@cse.psu.edu \vspace{0.3em}}}

\maketitle


\begin{abstract}
The purpose of binary segmentation models is to determine which pixels belong to an object of interest (e.g., which pixels in an image are part of roads). The models assign a logit score  (i.e., probability) to each pixel and these are converted into predictions by thresholding (i.e., each pixel with logit score $\geq \tau$ is predicted to be part of a road). However, a common phenomenon in current and former state-of-the-art segmentation models is spatial bias -- in some patches, the logit scores are consistently biased upwards and in others they are consistently biased downwards. These biases cause false positives and false negatives in the final predictions. In this paper, we propose PatchRefineNet (\ours), a small network that sits on top of a base segmentation model and learns to correct its patch-specific biases. Across a wide variety of base models, \ours consistently helps them improve mIoU by 2-3\%. One of the key ideas behind \ours is the addition of a novel supervision signal during training. Given the logit scores produced by the base segmentation model, each pixel is given a pseudo-label that is obtained by optimally thresholding the logit scores in each image patch.
Incorporating these pseudo-labels into the loss function of \ours helps correct systematic biases and reduce false positives/negatives. Although we mainly focus on binary segmentation, we also show how \ours can be extended to saliency detection and few-shot segmentation. We also discuss how the ideas can be extended to multiclass segmentation.

\end{abstract}
\section{Introduction}\label{sec:intro}
\begin{figure}[htpb]
    \centering
    \includegraphics[width=\linewidth]{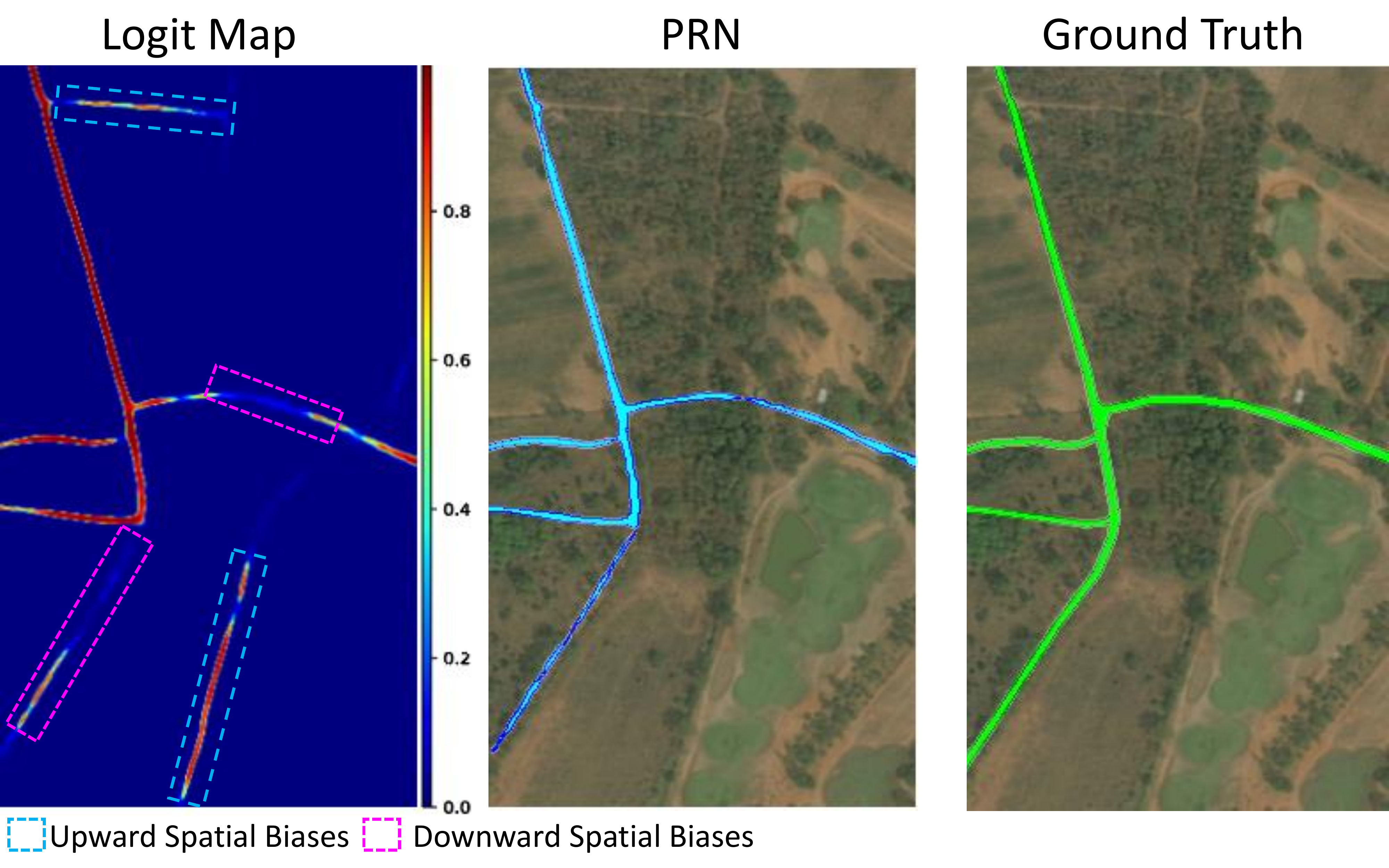 }
    \caption{\textbf{Spatial bias example in an image from DeepGlobe \cite{demir2018deepglobe}.} Column 1: Logit map produced by CoANet \cite{21TIP-Road}. Regions with significant \textcolor{magenta}{downward} and \textcolor{cyan}{upward} spatial biases  are highlighted by boxes with the corresponding colors.  Column 2: De-biased logit map produced by PRN. Column 3: Ground Truth.}
    \label{fig:road}
\end{figure}

Binary segmentation \cite{li2020fss, wang2017learning, demir2018deepglobe, jha2020kvasir} is the task of identifying which pixels in an image belong to objects of interest.  Examples include identifying roads in satellite images \cite{zhou2018d, zheng2020foreground, alirezaie2019semantic} and polyps in medical images \cite{pham2000survey, jha2020kvasir, zhou2018unet++}.
Neural networks that are trained to perform binary segmentation typically output something called a \emph{logit score} for each pixel -- a number between 0 and 1  indicating the likelihood that this pixel belongs to the object of interest. The logit scores for all the pixels are collectively referred to as a \emph{logit map}. The logit maps are converted into a final predictions through \emph{binarization} --picking a threshold $\threshold$ and setting a pixel's prediction to 1 if the corresponding logit is $\geq \threshold$ and 0 otherwise.

Despite steady improvement in  network architecture for binary segmentation models \cite{liu2015parsenet, badrinarayanan2017segnet, ronneberger2015u, chen2014semantic}, logit maps from former and current state-of-the-art networks exhibit spatial biases that limit the accuracy of the resulting binarized predictions. As an example, consider  \cref{fig:road}. The first column shows the logit map produced by CoANet \cite{21TIP-Road}, a high-performing road segmentation network, from an image from the DeepGlobe dataset \cite{demir2018deepglobe} (the  ground truth is shown in column 3). In the first column, the regions marked by \textcolor{magenta}{pink} boxes represent image patches with significant downward bias in their logit scores. In these patches, the pixels that actually belong to roads have an average logit score of  $\approx 0.2$. Meanwhile,  the \textcolor{cyan}{cyan} boxes represent image patches with significant upward bias in their logit scores. The lines shown inside those boxes have an average logit score of $\approx 0.7$. Having significant amounts of \emph{non-road} pixels with higher logit scores than actual road pixels is problematic --
binarization will produce final predictions with many false negatives in the \textcolor{magenta}{pink} boxes and false positives in the \textcolor{cyan}{cyan} boxes. This type of spatial bias in logit maps is not specific to CoANet -- it is a consistent trend for all segmentation networks we have tried. Meanwhile, the second column shows how our proposed PatchRefineNet (PRN) has removed the spatial biases.

Clearly, to handle spatial biases, the logit maps in different image patches should be handled differently (instead of being binarized in the same exact way). One naive approach is to allow each image patch to have its own threshold, and to have a neural network trained to predict what that patch-specific threshold should be (e.g., if it believes that logits are biased upward in an image patch, it can set a higher threshold for that patch). However, such an approach has an important shortcoming -- it is too rigid.
Even inside an image patch, there could be spatial variation in the bias. For example, in an image patch that is generally biased upwards, there will be many clusters of pixels with upwardly-biased logit scores, but there can still be clusters with downward biases or almost no biases. Binarizing such a patch with a single threshold can often result in clusters of false negatives/positives.

%

To address this problem, we propose \textbf{PatchRefineNet (PRN)}.\footnote{Source code and pre-trained models will be made available on GitHub upon acceptance.} One takes any segmentation network as a base and puts PRN on top of it (the input to PRN is the logit map produced by the base network). PRN learns the spatial biases of the base network and then adjusts the logit score of each pixel to compensate. PRN uses two learning signals during training. The first is the ground truth labeling of each pixel. The second is a novel learning signal from a set of ``pseudo-labels'' designed as follows:  (1) for each image patch in a training image, one first finds an optimal threshold for binarizing that patch; (2) then one uses these patch-specific thresholds to binarize each patch. The resulting binarization of each pixel is the pseudo-label for that pixel. Intuitively, these pseudo-labels train  PRN to detect the overall bias in a patch, while the ground truth learning signal trains PRN to detect the exceptions (e.g., clusters of pixels with a downward bias inside a patch that is generally upward-biased).

In order to learn about spatial biases in the base network, PRN splits an input logit map into $\numpatches$ disjoint patches. There is a global branch that processes the entire logit map, which helps PRN understand the relationships between patches. There is also a local branch that processes individual patches (to learn about local properties/biases in a patch). Both branches produce logit maps which are then averaged (resulting in the ``final'' logit map) and then thresholded at 0.5 (for final binarized predictions).

Why don't existing networks automatically correct their own biases by training with the ground truth? We conjecture this is because in their training, the loss at a pixel-only level depends on the label and prediction for the pixel, hence the networks are not very good at noticing general trends in their errors for clusters of pixels. On the other hand, the pseudo-labels used by PRN during training reflect collective trends in bias in different patches.

We train PRN separately from the base network for several reasons. The first reason is that if a trained base network already exists (e.g., a state-of-the-art from prior work), then this reduces resource (e.g., electricity) consumption compared to retraining everything from scratch. The next reason is that once the base network is fixed, its logit maps for each training image won't change. Hence PRN can avoid expensive re-computation of the pseudo-labels it needs.
Finally, the learning signal from the novel pseudo-labels  used by PRN does not have a meaningful derivative with respect to the weights of the base network -- the pseudo-labels are 0/1-valued numbers computed from the logit map of the base network; therefore the derivative with respect to the weights of the base network is either 0 or the delta function and hence does not work well with stochastic gradient descent-style optimization.


\noindent To summarize, our main contributions are:
\setlist{nolistsep}
\begin{itemize}[noitemsep,leftmargin=*]   
\item We propose PatchRefineNet (\ours), a post-processing network that sits on top of a base segmentation model and learns to correct its spatial biases.
\item PRN uses a novel learning signal that is computed from binarizing each patch separately and optimally. 
\item PRN complements virtually any binary segmentation network. In our experiments across different base models, \ours  consistently improves the mean Intersection over Union (mIoU) \cite{rezatofighi2019generalized} and mean Boundary Accuracy (mBA) \cite{cheng2021boundary} by 2-3\% over the base networks and hence there is good reason to believe that it can help future state-of-the-art networks improve their performance. 
\item We also explain how \ours can be extended to  saliency detection,  few-shot segmentation, and multi-class segmentation.

\end{itemize}

\section{Related Work}\label{sec:related_work}
\noindent\textbf{Semantic Segmentation Architectures.}\\ Previous methods for semantic segmentation \cite{jha2019resunet++, zhou2018d, zhao2019pyramid, li2020fss, nagendra2022constructing, 21TIP-Road} have been successful in extracting contextual information with wide fields-of-view \cite{chen2017deeplab, chen2016attention, farabet2012learning, he2004multiscale, nagendra2017comparison} along with FCN's \cite{long2015fully} bottom-up approach for better segmentation quality. This includes feature pyramid methods \cite{liu2015parsenet, he2015spatial,cheng2020cascadepsp, funk2018learning} that spatially pool \cite{liu2015parsenet, zhao2017pyramid} feature maps of different receptive fields, or dilated convolutions \cite{yu2015multi, chen2017deeplab, chen2018encoder, li2018pyramid} with different dilation rates. 
Encoder-decoder models \cite{badrinarayanan2017segnet, chen2018encoder, li2018pyramid, liu2019auto,liu2018path,noh2015learning,ronneberger2015u, liu2021new, pei2021utilizing, pei2020cloud} have been widely used in semantic segmentation. The encoder reduces spatial resolution to capture high-level global semantics, followed by a decoder which restores spatial resolution. Skip connections \cite{ronneberger2015u, cheng2020cascadepsp, nagendra2020efficient} can be further added to recover lost spatial information in deeper layers. Self-attention \cite{oktay2018attention, huang2019interlaced, ye2019cross, zhou2021self} has been used in segmentation networks to highlight salient features from context-rich skip connections and feature maps from deeper layers, where attention coefficients are more sensitive to local regions. Multi-scale context aggregation \cite{chen2014semantic, chen2016attention, xia2016zoom, hariharan2015hypercolumns} has proven to be efficient for integrating global and local features with two branches. Even though this alleviates higher memory usage arsing from using large output strides \cite{long2015fully, chen2017deeplab}, each branch has to be trained separately. 

\ours adopts the encoder-decoder architecture with skip connection and self-attention modules. Pyramid pooling is used for context aggregation at the bottleneck. \ours uses global and local branch decoders 
and allows for quick training and inference while being able to capture global and local structure from input logit maps.  
\\
\noindent\textbf{Segmentation Refinement.}
\\
FCN based methods typically do not generate very high-quality segmentation \cite{cheng2020cascadepsp}. State-of-the-art network architectures have modules that increase field-of-view for constructing reliable context information \cite{zhao2017pyramid, yuan2018ocnet, zhang2019co, fu2019dual, huang2019interlaced, chen2017deeplab, chen2016attention, farabet2012learning, he2004multiscale, nagendra2017comparison}, and/or increase resolution of feature maps \cite{chen2017rethinking, chen2018encoder, sun2019high} to achieve better segmentation performance. Separate boundary refinement modules \cite{zhang2019canet, peng2017large} are also used to improve boundary accuracy. They are typically large models trained in an end-to-end fashion. However, they have limited refinement capability \cite{cheng2020cascadepsp, zhang2019canet, chen2018encoder, yuan2020segfix} and inconsistencies \cite{chen2014semantic, yuan2020segfix} still exist in their final binarized predictions due to inherent spatial biases \cite{krygier2021quantifying, czolbe2021segmentation} in their output logit maps. Researchers have previously addressed the refinement process with postprocessing techniques like  Conditional Random Fields (CRF) \cite{chen2014semantic, chen2017deeplab, lin2016efficient, zheng2015conditional} or region growing \cite{dias2020probabilistic, dias2018semantic}. Other methods use cascading \cite{lin2017refinenet, cheng2020cascadepsp} and multi-scale context  aggregation \cite{chen2019collaborative, wu2020patch} to generate high resolution segmentation maps. These methods aim at coarse-fine iterative refinement.
However, to the best of our knowledge, no previous work has addressed refinement by focusing on correcting spatial biases \cite{krygier2021quantifying, czolbe2021segmentation} from raw logit maps other than \ours. In comparison with \ours, above mentioned graphical methods \cite{chen2014semantic, chen2017deeplab, lin2016efficient, zheng2015conditional, dias2020probabilistic, dias2018semantic} cannot fix large errors and they adhere to local semantics without fully leveraging global structure. Cascading and context aggregation methods \cite{lin2017refinenet, cheng2020cascadepsp} do not allow for single-stage training as individual patches are processed separately. Training time increases significantly with added levels of cascading. Additionally, as the patch size increases, memory usage also increases. On the contrary, \ours is a one-pass refinement module that allows for quick training and inference. Further, memory usage of \ours is constant for all patch sizes.

\section{The Patch Refine Network (\ours)}\label{sec:approach}
We next discuss the \ours architecture    (\cref{sec:framework}),  how psuedo-labels are generation during training (\cref{sec:gtgen}), and the loss function used for training (\cref{sec:loss}). While the main focus here is on binary segmentation, we also discuss how to extend \ours to multi-class segmentation (\cref{sec:multi}).

\subsection{Architecture}\label{sec:framework}
Given a trained binary segmentation model that serves as a base,
\ours is designed to be a small, lightweight network that sits on top of this base and learns to correct its spatial biases. Its input is a logit map (produced by running an image through the base network) and its output is a logit map that can be thresholded at $0.5$ to create binary predictions. 
\begin{figure}[htpb]
    \centering
    \includegraphics[width=1.08\linewidth]{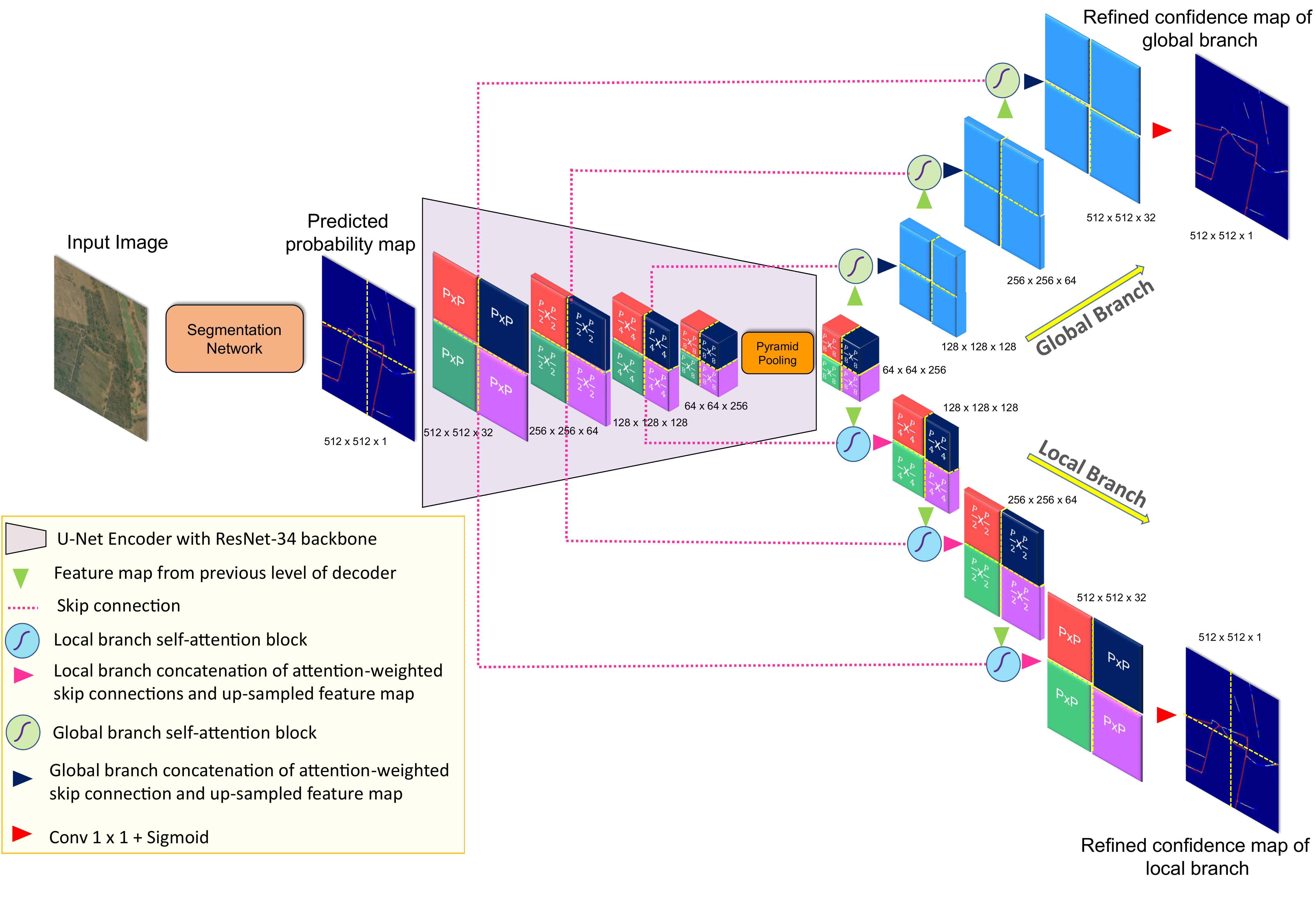}
    \caption{\textbf{\ours framework. }Example of \ours architecture configured for $k=4$ patches. \ours has a U-Net \cite{ronneberger2015u} encoder with ResNet-34 \cite{he2016deep} backbone with $\frac{1}{8}$ resolution scaling and pyramid pooling at the bottleneck. There are two decoder branches -- a \emph{global} and \emph{local}. 
    }
    \label{fig:figure4}
\end{figure}

It is common practice for such networks, which post-process the output of a base network \cite{chen2019collaborative, wu2020patch}, to have both a \emph{global} branch to capture the overall structure in an input and a \emph{local} branch to analyze finer structures.
%
%
%
\ours follows a similar strategy. While prior post-processing networks had to train local and global branches separately, \ours is able to jointly train the encoder and local/global decoders. The local and global branches both produce a logit map and the two maps are averaged during inference.

First, the logit map produced by the base segmentation network is resized to $512 \times 512$ and given as input to the \ours encoder. The output of the encoder, referred to as the \textit{bottleneck}, is a feature map  that has $\frac{1}{8}$ resolution scaling. This is simultaneously passed through the global and local decoder branches. The global decoder takes the entire feature map at the bottleneck as input. 

Before running the local branch, the output of the encoder is split into $k$ disjoint patches. The value of $k$  is determined by a patch-size parameter $P$ as follows: $k=(512/P)^2$. Each patch is independently sent through the local branch to produce $1/k^\text{th}$ of the logit map. The full logit map of the local branch is then re-assembled from these pieces after the local branch processes all patches.
%

\indent \textbf{Encoder:}  We use a standard U-Net \cite{ronneberger2015u} encoder with ResNet-34 \cite{he2016deep} backbone to extract features from the input logit map, as shown in \cref{fig:figure4}. The spatial resolution decreases from $512$ to $64$ ($\frac{1}{8}^{th}$), while the number of features increases from $32$ to $256$ at the end of four encoding levels. After the fourth encoding level, pyramid pooling \cite{zhao2017pyramid} with pooling sizes [1, 2, 4, 8] is used for rich global contextual features. The final resolution at the bottleneck, after pyramid pooling, is $64 \times 64 \times 256$. 


\indent\textbf{Global Decoder Branch:} The core purpose of this decoder is to capture global inter-patch semantics. This is a typical U-Net decoder with convolutional blocks, as shown in  \cref{fig:figure4}. We add a self-attention \cite{oktay2018attention} layer at each decoding level before concatenating the skip connection from the previous encoder level with an upsampled feature map from the previous decoder layer. Self-attention filters highlight salient features from spatial-information-rich skip connections and context-rich decoder (deeper) layers. This branch is used to extract the relationship between patches. \\
\indent\textbf{Local Decoder Branch:}
\begin{figure}[htpb]
    \centering
    \includegraphics[width=\linewidth]{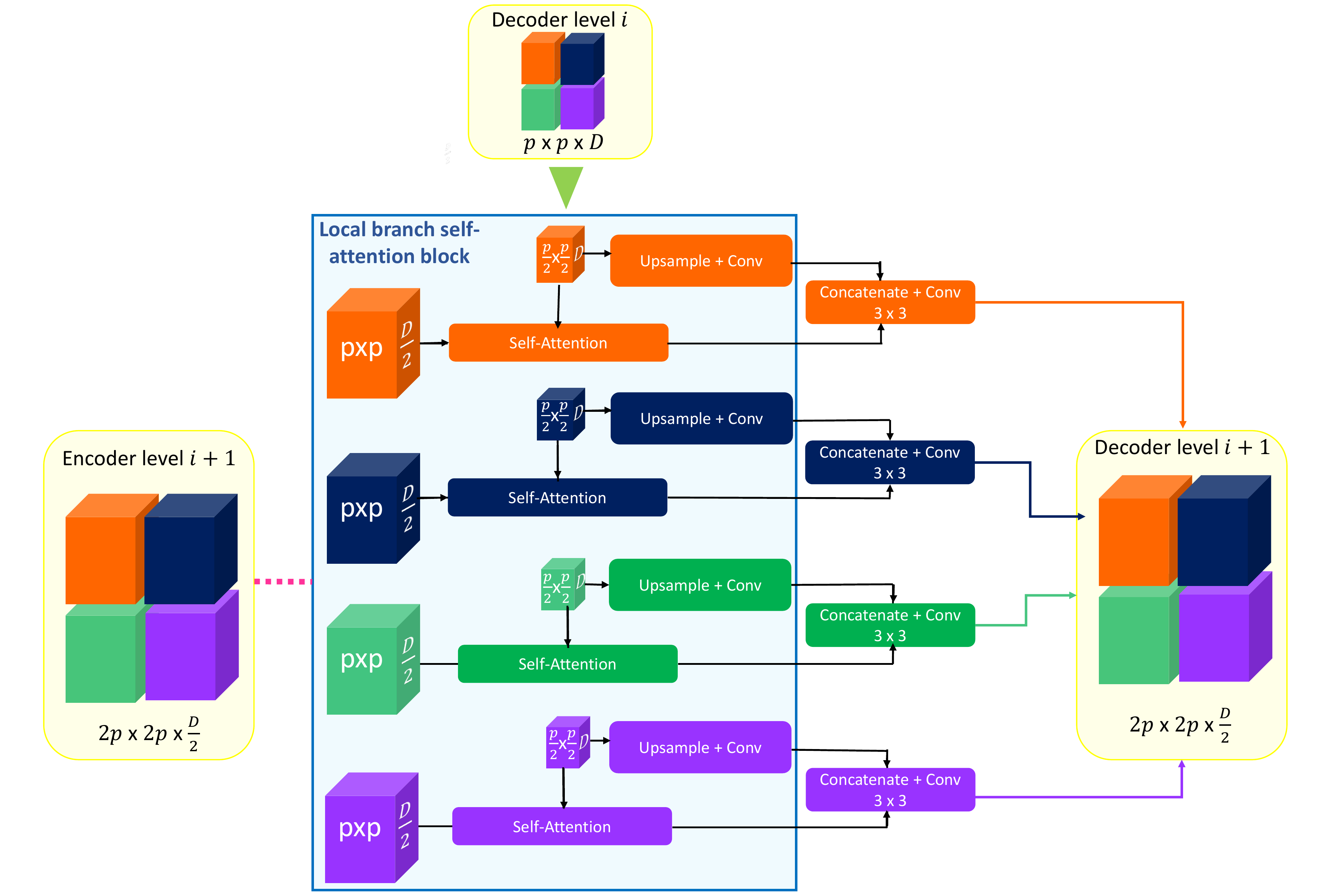}
    \caption{\textbf{Local branch self-attention block}, configured for $k=4$  patches ($P = 256$). Inputs to the block are the feature map from decoding level $i$ and skip connection from encoding level $i+1$.}
    \label{fig:figure5}
\end{figure}
The core purpose of this decoder is to capture local intra-patch semantics from each feature patch. A magnified version of the local branch self-attention block is shown in \cref{fig:figure5}. The feature map from decoding level $i$ has size $p\times p \times D$, where $p = \frac{P}{2^i}$, and  $D$ is the number of filters at decoding level $i$. It is broken down into $k$ patches of size $\frac{p}{2} \times \frac{p}{2} \times D$. The skip connection from encoding level $i+1$ has size $2p \times 2p \times \frac{D}{2}$ and is also broken down into $k$ patches of size $p\times p \times \frac{D}{2}$. Each patch from decoding level $i$ is upsampled and concatenated with the attention-weighted patch from the encoding level $i+1$. Finally, processed patches are spatially merged to size $2p \times 2p \times \frac{D}{2}$, which is the output of decoding level $i+1$. This helps in capturing local semantics from each patch. The design of this branch allows the training to be performed simultaneously, as opposed to other patch aggregation and cascading methods\cite{chen2019collaborative, wu2020patch}.\\\indent\textbf{Inference:}
During inference, the logit map output by the base network is passed as the input to the \ours. Both the local and global branches produce logit maps which are then averaged (resulting in the “final” logit map) and then thresholded at 0.5 (for final binarized predictions).

\subsection{Generating Patch-Optimal Thresholded Maps}\label{sec:gtgen} 
The data used for tuning the base network's hyperparameters also serves to train \ours. One  learning signal (used in the loss function in Section \ref{sec:loss}) is the ground truth labeling $Y$ of an image.  The other is a novel set of ``pseudo-labels''. Let $\widehat{Y}$ be the logit map produced by the base network.
This $\widehat{Y}$ is split into $k$ patches. For each patch $p_i$, one finds the threshold that maximizes the mIoU for that patch (when the patch is binarized using the threshold). The resulting binarized patches are the pseudo-labels.

The intuition behind the pseudo-labels is that the most efficient way to minimize  loss (e.g., binary cross-entry) between an output logit map and 
the pseudo-labels is to shift an entire patch from a logit map up or down. For example, in the case of the very last layer this is achieved by mainly focusing on the bias parameter of the layer. Thus pseudo-labels lets the network focus on aggregate properties (e.g., spatial biases) of the patch, whereas the ground truth signal makes the network focus on properties of individual pixels.


\subsection{Loss Function}\label{sec:loss}
\noindent 
The loss function uses the two learning signals defined above. The global and local branches both use the same loss function, and the overall loss is the sum of the two. Hence we describe the loss $\loss$ for one of the branches.

$\loss$ is the weighted sum of two components, a loss $\loss_{gt}$ with respect to the ground truth and a loss $\loss_{ps}$ with respect to the pseudo-labels:
\begin{equation}
    \loss = \alpha {\loss_{ps}} + (1 - \alpha){\loss_{gt}} \quad \text{with } \alpha = 0.7,\label{eq:totalloss}
\end{equation}
where $\alpha$ was tuned based on 100 randomly augmented images from the DeepGlobe training set \cite{demir2018deepglobe}.

$\loss_{gt}$ is the standard binary cross-entropy loss \cite{zhang2018generalized} between the ground truth and the logit map produced by the branch.


 ${\loss_{ps}}$ uses the pseudo-labels for the ground truth and can be written as a sum: ${\loss_{ps}} = \loss_{focal} + \loss_{boundary}$,
where $\loss_{focal}$ is known as the \emph{focal loss} \cite{lin2017focal} and $\loss_{boundary}$ is known as the \emph{boundary loss} \cite{zhao2019pyramid}. Both focal and boundary loss are standard in image segmentation, however we use the pseudo-labels, generated for patch-size parameter $P$, in place of the ground truth in the computation of the losses. Focal loss \cite{lin2017focal} is a variation of binary cross-entropy loss  that introduces a parameter $\gamma$ (tuned using the same 100 DeepGlobe images as $\alpha$ in Equation \ref{eq:totalloss}). For a pixel $i$, let $c_i$ be the output of the branch for that pixel (i.e., a logit value) and let $\widetilde{y}_i$ be the pseudo-label. Then 
\begin{align*}
\loss_{focal}=-\sum_{i~:~\widetilde{y}_i=1} (1-c_i)^\gamma\log(c_i)-\sum_{i~:~\widetilde{y}_i=0} c_i^\gamma\log(1-c_i)
\end{align*}

Boundary loss \cite{zhao2019pyramid} is designed to improve predictions at boundary pixels. It is computed as follows. Let $C$ be the matrix corresponding to the logit map output by a branch. The squashed Laplace operator \cite{zhao2019pyramid} applied to $C$ is:
\begin{align*}
\abs(\tanh(\conv(C, K))\quad\text{ where }k=\left(\begin{smallmatrix}
0 & 1 & 0\\
1 & -4 & 1\\
0 & 1 & 0\\
\end{smallmatrix}\right)
\end{align*}
The boundary loss $\loss_{{boundary}}$ \cite{zhao2019pyramid}
is the defined as the binary cross entropy between the squashed Laplace operator applied to $C$ and the squashed Laplace operator applied to the target labels, which in our case are the pseudo-labels.

%
%

\subsection{Extension to Multi-Class Segmentation}\label{sec:multi}
In this section, we explain how this technique could be extended to multiclass semantic segmentation with $\numclasses$ classes. The output at each pixel, instead of being a single logit, is now a $\numclasses$-dimensional vector produced by the softmax activation. If we let $\vec{x}_i$ denote the pre-activation at pixel $i$, then the output at the pixel is $\softmax(\vec{x}_i)$. 

The pseudo-label for a pixel becomes a $\numclasses$-dimensional one-hot encoding vector. During training it can be generated as follows. Previously, the best threshold was used to binarize each image patch. In the multiclass setting, the threshold is replaced by a $\numclasses$-dimensional vector $\vec{t}$ and the ``pseudo-label class'' for a pixel is chosen by the formula:  $\arg\max_{j} (\vec{t}[j] + \softmax(\vec{x}_i)[j])$ -- this is the same as translating the softmax by the vector $\vec{t}$ and choosing  class $j$ if the $j^\text{th}$ component is the largest. The pseudo-label is the one-hot encoding of the chosen class.

The difficulty here is in choosing the optimal $\vec{t}$ for each image patch in the training data.
In the binary case, we were only dealing with a threshold, and it was easy to try different numbers between $0$ and $1$. However, this becomes inefficient when searching for the optimal vector $\vec{t}$. Performing this search efficiently is part of our future work, and our goal in this paper is to evaluate how well \ours works in the binary segmentation setting.


\section{Experiments}\label{sec:results}
In this section, we evaluate the ability of \ours to improve the prediction of a base binary segmentation network. We consider a variety of datasets and base networks (including current and former state-of-the-art segmentation models) along with other postprocessing methods. Overall, \ours consistently improves performance in mIoU by approximately 2-3\% and thus is likely to help future models improve their predictions as well, by reducing their spatial biases.

\subsection{Datasets}\label{sec:data}
We use the following four datasets for evaluation: DeepGlobe \cite{demir2018deepglobe}, Kvasir-SEG \cite{jha2020kvasir}, DUTS \cite{wang2017learning}, and FSS-1000 \cite{li2020fss} on three types of tasks: binary segmentation (DeepGlobe, Kvasir-SEG), saliency detection (DUTS), and few-shot segmentation (FSS-1000).

DeepGlobe \cite{demir2018deepglobe} is a large-scale road extraction dataset that contains 6226 labeled images. We divide this into 4980 training images, 996 validation images, and 250 test images. Kvasir-SEG \cite{jha2020kvasir} is a large-scale polyp segmentation dataset with 1000 labeled images. 
DUTS \cite{wang2017learning} contains 10553 images for training and 5019 images for evaluation. We divide these 5019 images into 4015 validation images and 1004 test images. FSS-1000 \cite{li2020fss} contains 1000 classes with 10 images each. We divide the 1000 classes into 760 classes for training, 192 classes for validation, and 48 classes for testing. Each class contains 10 images out of which we use 5 images as support (labeled images to generalize from for few-shot learning) and the other 5 as query (test images). 

\subsection{Evaluation Criteria}
Similar to prior work in binary segmentation, we use mean Intersection over Union \cite{rezatofighi2019generalized} (mIoU) and mean Boundary Accuracy (mBA)  \cite{cheng2021boundary} as the evaluation metrics. mBA, also called boundary mIoU, is a new measure proposed by \cite{cheng2021boundary} which has a weaker bias toward large objects than mIoU. It neither over-penalizes nor ignores errors in small objects. Given the matrix of ground truth  pixel labels and (binarized) predicted labels, Boundary mIoU first computes the set of the pixels that are within a distance $d$ from each contour (computed from \cite{opencvOpenCVContours}) in the ground truth  and in the predictions and then computes mIoU of these two sets.  We use $d=15$ as recommended in \cite{cheng2021boundary}.  We evaluate the performance of \ours on Saliency detection \cite{zhao2019pyramid} using mean absolute error (MAE), along with mIoU and mBA.

\subsection{Implementation Details}
The base networks are trained according to the code and  implementation details provided in the respective papers.

The datasets we use are divided into training, validation, and test sets as discussed in \cref{sec:data}. The train set is used to train the base model. The validation set is used to tune the hyperparameters of the base model and to train \ours (the validation set is never used for reporting). To make sure comparisons are fair, we also try settings where the base model include the validation data in training (\cref{sec:exp:valid}). Finally, the test set is used for reporting results.


Data augmentations such as random rotation, and horizontal and vertical flips are used for training the models. \ours is trained with the Adam \cite{kingma2014adam} optimizer with an initial learning rate of $8e^{-4}$, batch size of 4, and for a maximum of 300 epochs on an NVIDIA 2080 Ti GPU. The learning rate is decreased until $5e^{-8}$. We use early stopping if its training loss does not decrease for 10 epochs.

\subsection{Ablation Experiments}\label{sec:ablation}
We first present ablation studies using the DeepGlobe \cite{demir2018deepglobe} dataset and base network DLinkNet \cite{zhou2018d}.

\subsubsection{Role of the validation set.} \label{sec:exp:valid}
Ordinarily, the base model would train on the training set and tune hyperparameters  on the validation set, which is also used to train \ours (we emphasise that \emph{results} are reported on the \emph{test} set only, which is disjoint from validation and train).  \ours uses the validation data because this is where the spatial bias of the base models become apparent.

This raises the question of whether it is a fair setup -- would it be better to simply add the validation data to the base model's training set and not use \ours? To answer this question, we consider the following 3 cases. (\textbf{A}) The base network trains on training data and tunes hyperparameters on validation data; \ours is not used. (\textbf{B}): The base network is trained using the combined training and validation data; we use the default hyperparameters from the DLinkNet github repository \cite{githubGitHubZlckanataDeepGlobeRoadExtractionChallenge}; \ours is not used. (\textbf{C}): The base network trains on training data and tunes hyperparameters on validation data; \ours is then trained on the validation data. The results, reported on the test set (disjoint from train and validation) are shown in Table \ref{tab:val}.


\begin{table}[htpb]
\centering
\begin{tabular}{l|cc}
\hline
\multirow{2}{*}{\textbf{Experiments}} & \multicolumn{2}{c}{\textbf{DeepGlobe \cite{demir2018deepglobe} test-set}}    \\  
                                      & \multicolumn{1}{c|}{mIoU (\%)} & mBA (\%) \\ \hhline{===}
\begin{tabular}{l}
\textbf{A}: Train on train set, tune \\
 on validation, no \ours
\end{tabular}
& \multicolumn{1}{c|}{61.3}      & 49.8     \\ \hline
\begin{tabular}{l}
\textbf{B}: Train on train and\\ validation set, no \ours
\end{tabular}
& \multicolumn{1}{c|}{59.7}      & 48.4     \\ \hline
\begin{tabular}{l}
\textbf{C}: Train on train set, tune \\
 on validation, yes \ours
\end{tabular}                          & \multicolumn{1}{c|}{\textbf{64.4}}      & \textbf{56.6}     \\ \hline
\end{tabular}
\caption{Evaluating the role of the validation set. }
\label{tab:val}
\end{table}

As we can see, reserving some data for hyper-parameter tuning is beneficial to the base network (case A improves upon case B). Re-using this validation set to train \ours shows a further, significant boost (case C is by far the best). This validates our proposed setup for how different parts of the data are used.

\subsubsection{Choosing  patch size parameter $P$}

\begin{figure}[htpb]
    \centering
    \includegraphics[width=\linewidth]{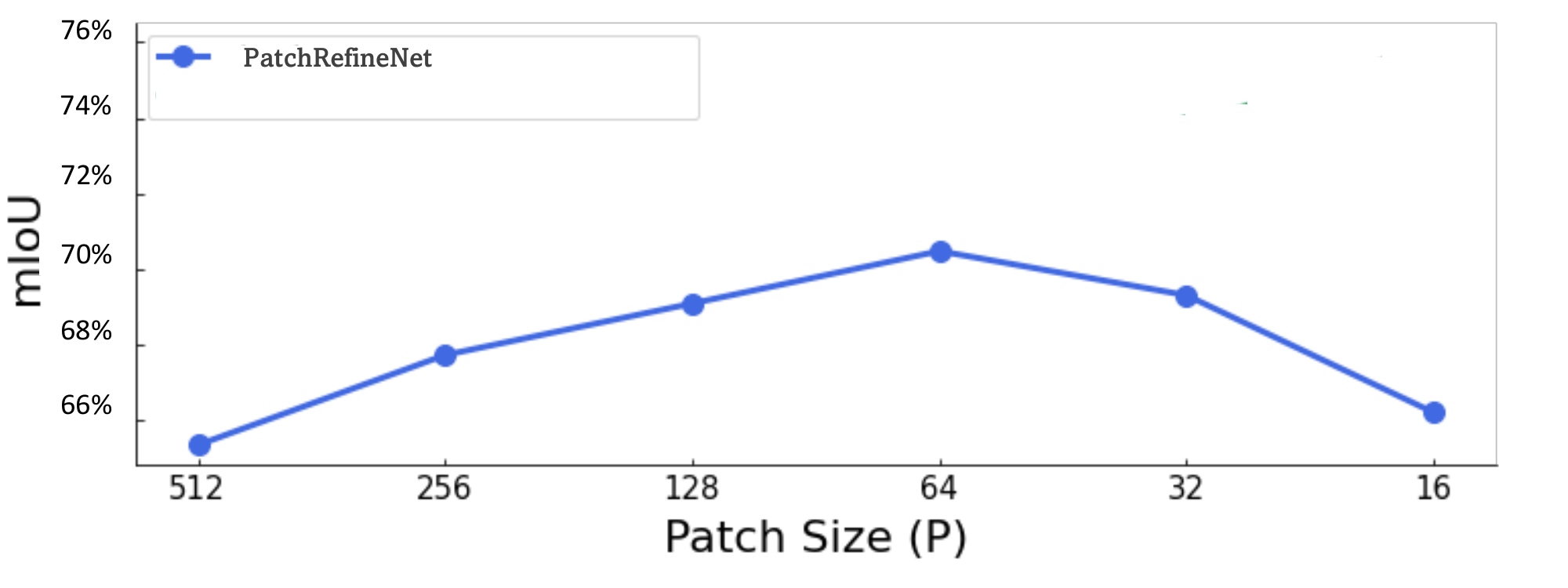}
    \caption{Effect of patch size parameter $\mathbf{P}$. mIoU is computed using  the  DeepGlobe \underline{training} set. 
    }
    \label{fig:patch}
\end{figure}

\noindent The patch size used by the local branch of \ours is controlled by the parameter $P$.  \cref{fig:patch} shows  the effect of $P$ on mIoU, which is evaluated on 100 randomly augmented images from the DeepGlobe \underline{training} set. The best choice, $P=64$, results in the local branch dividing the input logit map into sixty-four patches of size $64\times 64$. Since the testing set was not used at all for choosing patch size, it is appropriate to use $P=64$ in the rest of our experiments.


\subsubsection{Ablation study of \ours design.} 
We next consider an ablation study of the rest of the design of \ours, including the benefit of using of global/local branches and a loss function based on pseudo-labels.

It is becoming increasingly common to use global and local branches to improve segmentation quality \cite{chen2019collaborative, wu2020patch}. In the case of \ours, where we want to detect and correct patch-specific spatial biases, local branches are clearly necessary from the design perspective. \begin{table}[htpb]
\centering
\resizebox{\columnwidth}{!}{%
\begin{tabular}{lcc}
\multicolumn{1}{c|}{\multirow{2}{*}{\textbf{Configuration}}} & \multicolumn{2}{c}{\textbf{DeepGlobe \cite{zhou2018d}  test-set}} \\
\multicolumn{1}{c|}{}                                        & \multicolumn{1}{c|}{mIoU (\%)}       & mBA (\%)        \\ \hhline{===}
\multicolumn{1}{c|}{\textbf{\begin{tabular}[c]{@{}c@{}}Base Network: D-LinkNet \cite{demir2018deepglobe} \end{tabular}}} & \multicolumn{1}{c|}{61.3} & 49.8 \\ \hline
\multicolumn{3}{c}{\textbf{Ablation of network design}}                                                               \\ \hline
\multicolumn{1}{l|}{Global branch only}                      & \multicolumn{1}{c|}{$61.7_{\uparrow 0.4}$}            & $52.3_{\uparrow 2.5}$            \\
\multicolumn{1}{l|}{Local branch only}                       & \multicolumn{1}{c|}{$63.5_{\uparrow 2.2}$}            & $56.1_{\uparrow 6.3}$           \\
\multicolumn{1}{l|}{Local + Global (ours)}                   & \multicolumn{1}{c|}{$\mathbf{64.4_{\uparrow 3.1}}$}   & $\mathbf{56.6_{\uparrow 6.8}}$   \\ \hline
\multicolumn{3}{c}{\textbf{Ablation of total loss function}}                                                                \\ \hline
\multicolumn{1}{l|}{$\mathbf{L}_{gt}$ only}                          & \multicolumn{1}{c|}{$62.0_{\uparrow 0.7}$}            & $52.8_{\uparrow 3.0}$            \\
\multicolumn{1}{l|}{$\mathbf{L}_{ps}$ only}               & \multicolumn{1}{c|}{$63.8_{\uparrow 2.5}$ }           & $57.2_{\uparrow 7.4}$           \\
\multicolumn{1}{l|}{$\mathbf{L}_{gt}$ + $\mathbf{L}_{ps}$ (ours)} & \multicolumn{1}{c|}{$\mathbf{64.4_{\uparrow 3.1}}$}   & $\mathbf{56.6_{\uparrow 6.8}}$   \\ \hline
\multicolumn{3}{c}{\textbf{Ablation of Region-specific loss $\mathbf{L_{ps}}$}}                                                                \\ \hline
\multicolumn{1}{l|}{$\loss_{focal}$ only}                          & \multicolumn{1}{c|}{$64.0_{\uparrow 2.7}$}            & $51.9_{\uparrow 2.1}$            \\
\multicolumn{1}{l|}{$\loss_{boundary}$ only}               & \multicolumn{1}{c|}{$62.2_{\uparrow 0.9}$ }           & $55.4_{\uparrow 5.6}$ \\
\multicolumn{1}{l|}{$\loss_{focal} + \loss_{boundary}$ (ours)} & \multicolumn{1}{c|}{$\mathbf{64.4_{\uparrow 3.1}}$}   & $\mathbf{56.6_{\uparrow 6.8}}$   \\ \hline
\end{tabular}
}
\caption{Ablation results  for the design of \ours ($P=64$). 
}
\label{tab:ablation}
\end{table}
At the top of Table \ref{tab:ablation}, we compare performance when \ours includes a global branch only, local branch only, and both branches together.
As expected, the local branch is much more important than the global branch, with roughly a 2\% better mIoU and 4\% better mean boundary accuracy. Also, as expected, there is a very slight performance boost when the global branch is added to the local  branch, as this allows \ours to incorporate wider context information from the global branch.

Now, recall that the loss function in each branch is a sum of two losses $\loss_{gt}$ whose learning signal comes from the ground truth and $\loss_{ps}$ which comes from our pseudo-labels. The middle section of Table \ref{tab:ablation} shows the results of using only the ground truth ($\loss_{gt}$), only the pseudo-labels ($\loss_{ps}$), or both ($\loss_{gt}+\loss_{ps}$). Again we see that the pseudo-labels are more important than using the ground truth, probably because the base network is already trained with the ground truth signal, while the pseudo-labels summarize new information about systematic biases (as explained in \cref{sec:gtgen}). As expected, combining the two losses leads to a slight improvement over using pseudo-labels alone since the ground truth does contain information not present in pseudo-labels.

Finally, the loss over pseudo-labels, which is designed to correct patch-wise spatial biases is a mixture of focal loss \cite{lin2017focal} and boundary loss \cite{zhao2019pyramid}. Both are used in the literature to improve segmentation on fine structures, with boundary loss focusing on the boundary. As we can tell from the bottom of Table \ref{tab:ablation}, focal loss is better at improving mIoU while boundary loss is better at improving mean boundary accuracy, which is consistent with prior work. The combination of the two losses gives us the best of both worlds.

\subsection{Performance Evaluation } 
\subsubsection{Binary segmentation.} 
We next evaluate the improvement that \ours provides when combined with a variety of state-of-the-art and former state-of-the-art networks for binary segmentation on the DeepGlobe and Kvasir-SEG datasets. 
Table \ref{tab:results1}  shows that \ours provides consistent improvement by at least 2.3\% in mIoU and 2.6\% mBA  on both datasets for all networks, illustrating that they all have spatial bias, which \ours addresses. This supports the hypothesis that \ours is likely to help future networks to further improve their performance.
\begin{table}[htbp]
\centering
\begin{tabular}{ccc}
\hline
\multicolumn{3}{c}{\textbf{DeepGlobe \cite{demir2018deepglobe} test-set}} \\ \hhline{===}
\multicolumn{1}{c|}{\textbf{Methods}} &
  \multicolumn{1}{c|}{mIoU (\%)} &
  mBA (\%) \\ \hline
\multicolumn{1}{l|}{\begin{tabular}[c]{@{}l@{}}U-Net\cite{ronneberger2015u}\\ (+) \ours\end{tabular}} &
  \multicolumn{1}{c|}{\begin{tabular}[c]{@{}c@{}}55.8\\ $\mathbf{60.9_{\uparrow 5.1}}$\end{tabular}} &
  \begin{tabular}[c]{@{}c@{}}37.6\\ $\mathbf{47.4_{\uparrow 9.8}}$\end{tabular} \\ \hline
\multicolumn{1}{l|}{\begin{tabular}[c]{@{}l@{}}DeepLabV3+\cite{chen2017deeplab}\\ (+) \ours\end{tabular}} &
  \multicolumn{1}{c|}{\begin{tabular}[c]{@{}c@{}}59.2\\ $\mathbf{61.9_{\uparrow 2.7}}$\end{tabular}} &
  \begin{tabular}[c]{@{}c@{}}47.6\\ $\mathbf{55.9_{\uparrow 8.3}}$\end{tabular} \\ \hline
\multicolumn{1}{l|}{\begin{tabular}[c]{@{}l@{}}PSPNet \cite{zhao2017pyramid}\\ (+) \ours\end{tabular}} &
  \multicolumn{1}{c|}{\begin{tabular}[c]{@{}c@{}}59.8\\ $\mathbf{62.4_{\uparrow 2.6}}$\end{tabular}} &
  \begin{tabular}[c]{@{}c@{}}48.2\\ $\mathbf{56.6_{\uparrow 8.4}}$\end{tabular} \\ \hline
\multicolumn{1}{l|}{\begin{tabular}[c]{@{}l@{}}D-LinkNet \cite{zhou2018d} \\ (+) \ours\end{tabular}} &
  \multicolumn{1}{c|}{\begin{tabular}[c]{@{}c@{}}61.3\\ $\mathbf{64.4_{\uparrow 3.1}}$\end{tabular}} &
  \begin{tabular}[c]{@{}c@{}}49.8\\ $\mathbf{56.6_{\uparrow 6.8}}$\end{tabular} \\ \hline
\multicolumn{1}{l|}{\begin{tabular}[c]{@{}l@{}}GLNet \cite{chen2019collaborative} \\ (+) \ours\end{tabular}} &
  \multicolumn{1}{c|}{\begin{tabular}[c]{@{}c@{}}62.8\\ $\mathbf{65.4_{\uparrow 2.6}}$\end{tabular}} &
  \begin{tabular}[c]{@{}c@{}}52.6\\ $\mathbf{57.9_{\uparrow 5.3}}$\end{tabular} \\ \hline
\multicolumn{1}{l|}{\begin{tabular}[c]{@{}l@{}}ISDNet \cite{guo2022isdnet} \\ (+) \ours\end{tabular}} &
  \multicolumn{1}{c|}{\begin{tabular}[c]{@{}c@{}}64.8\\ $\mathbf{67.3_{\uparrow 2.5}}$\end{tabular}} &
  \begin{tabular}[c]{@{}c@{}}54.8\\ $\mathbf{59.2_{\uparrow 4.4}}$\end{tabular} \\ \hline
  \multicolumn{1}{l|}{\begin{tabular}[c]{@{}l@{}}CoANet \cite{21TIP-Road} \\ (+) \ours\end{tabular}} &
  
  \multicolumn{1}{c|}{\begin{tabular}[c]{@{}c@{}}67.9\\ $\mathbf{70.6_{\uparrow 2.7}}$\end{tabular}} &
  \begin{tabular}[c]{@{}c@{}}58.4\\ $\mathbf{62.1_{\uparrow 3.7}}$\end{tabular} \\ \hline
  
\multicolumn{3}{c}{\textbf{Kvasir-SEG \cite{jha2020kvasir} test-set}}                                  \\ \hhline{===}
\multicolumn{1}{l|}{\begin{tabular}[c]{@{}l@{}}U-Net \cite{ronneberger2015u}\\ (+) \ours\end{tabular}} &
  \multicolumn{1}{c|}{\begin{tabular}[c]{@{}c@{}}41.5\\ $\mathbf{47.8_{\uparrow 6.3}}$\end{tabular}} &
  \begin{tabular}[c]{@{}c@{}}38.8\\$\mathbf{46.3_{\uparrow 7.5}}$\end{tabular} \\\hline
\multicolumn{1}{l|}{\begin{tabular}[c]{@{}l@{}}ResUnet \cite{diakogiannis2020resunet}\\ (+) \ours\end{tabular}} &
  \multicolumn{1}{c|}{\begin{tabular}[c]{@{}c@{}}46.8\\ $\mathbf{52.9_{\uparrow 6.1}}$\end{tabular}} &
  \begin{tabular}[c]{@{}c@{}}45.7\\ $\mathbf{52.5_{\uparrow 6.8}}$\end{tabular} \\ \hline
\multicolumn{1}{l|}{\begin{tabular}[c]{@{}l@{}}ResUnet++ \cite{jha2019resunet++}\\ (+) \ours\end{tabular}} &
  \multicolumn{1}{c|}{\begin{tabular}[c]{@{}c@{}}55.9\\ $\mathbf{61.7_{\uparrow 5.8}}$\end{tabular}} &
  \begin{tabular}[c]{@{}c@{}}56.8\\ $\mathbf{62.9_{\uparrow 6.1}}$\end{tabular} \\ \hline
\multicolumn{1}{l|}{\begin{tabular}[c]{@{}l@{}}SSFormer-S \cite{wang2022stepwise}\\ (+) \ours\end{tabular}} &
  \multicolumn{1}{c|}{\begin{tabular}[c]{@{}c@{}}86.8\\ $\mathbf{89.1_{\uparrow 2.3}}$\end{tabular}} &
  \begin{tabular}[c]{@{}c@{}}69.7\\ $\mathbf{72.3_{\uparrow 2.6}}$\end{tabular} \\ \hline
\end{tabular}
\caption{How \ours helps base networks for binary segmentation.}
\label{tab:results1}
\end{table}
\begin{figure}[htpb]
    \centering
    \includegraphics[width=0.9\linewidth]{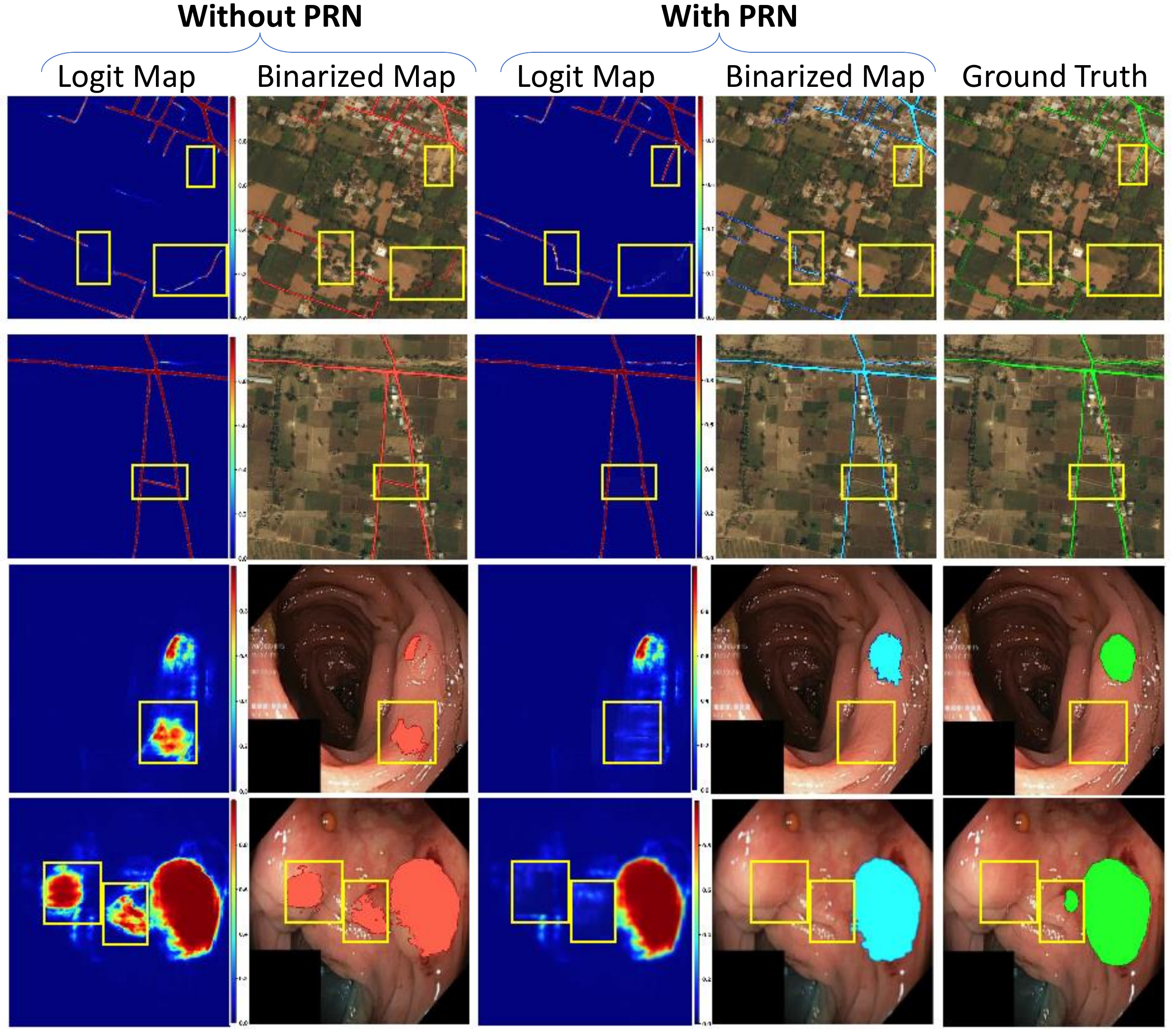}
    \caption{Qualitative examples of improvement due to \ours. Rows 1 \& 2: images from DeepGlobe with CoANet as the base network. Rows 3 \& 4: images from Kvasir with SSFormer-S as the base network. Ground truth: last column. Yellow boxes represent areas where \ours causes most improvement. Column 1: logit map of the base network. Column 2: binarized predictions of base network. Column 3: logit map output by \ours. Column 4: binarized predictions from \ours. 
    }    \label{fig:qual_deepglobe}
\end{figure}
 \cref{fig:qual_deepglobe} shows qualitative examples. The first two rows come from the DeepGlobe test set with CoANet \cite{21TIP-Road} as the base network; the task is to identify roads in the image. The last two rows are from the Kvasir-SEG test data with SSFormer-S \cite{wang2022stepwise} as the base; the task is to identify polyps. The first two columns show the logit map and binarized prediction, respectively, of the base network. The yellow boxes highlight areas of false positives and false negatives. The next two columns show the logit map and binarized prediction after \ours de-biases the base networks. The last column shows the ground truth. For example, in the first row, the left-most yellow box identifies a region where the base network missed part of a road, resulting in two disconnected road segments; this is a negative bias in that region that \ours fixes.
 In the second row, the base network predicts that the roads have an `A' shape but the cross-bar is a false positive that gets removed by \ours. The corrections made by \ours are more clearly visible in the last two rows.
%
%
%
\subsubsection{Comparison  with other post-processing methods}
Although the literature on post-processing methods is very sparse, DenseCRF \cite{zheng2015conditional} and CascadePSP \cite{cheng2020cascadepsp} are two notable postprocessing techniques for improving binary segmentation. 
Our first comparison, to DenseCRF, shows that \ours is much better at improving both mIoU and mBA. Due to space restrictions, a small subset of our results in shown in Table \ref{tab:crf}. More extensive comparisons with DenseCRF for all datasets can be found in the supplementary material.
\begin{table}[htpb]
\centering
\begin{tabular}{lcc}
\hline
\multicolumn{3}{c}{\textbf{DeepGlobe \cite{demir2018deepglobe} test-set}} \\ \hhline{===}
\multicolumn{1}{c|}{\textbf{Methods}} &
  \multicolumn{1}{c|}{mIoU (\%)} &
  mBA (\%) \\ \hline
\multicolumn{1}{l|}{\begin{tabular}[c]{@{}l@{}}CoANet\cite{21TIP-Road}\\ (+) DenseCRF \cite{zheng2015conditional}\\ (+) \ours\end{tabular}} &
  \multicolumn{1}{c|}{\begin{tabular}[c]{@{}c@{}}67.9\\ $69.0_{\uparrow1.1}$\\ $\mathbf{70.6_{\uparrow2.7}}$ \end{tabular}} &
  \begin{tabular}[c]{@{}c@{}}58.4\\ $59.6_{\uparrow1.2}$\\ $\mathbf{62.1_{\uparrow3.7}}$\end{tabular} \\ \hline
\end{tabular}
  \caption{Comparison to DenseCRF \cite{zheng2015conditional} postprocessing.}.
  \label{tab:crf}
\end{table}

CascadePSP \cite{cheng2020cascadepsp} is another post-processing technique that supports several different configurations, such as number of cascade levels in the global step and different image crop sizes. In Table \ref{tab:cascade}, we compare \ours with CascadePSP with different configurations. 
\begin{table}[htpb]
\centering
\resizebox{\columnwidth}{!}{%
\begin{tabular}{ccc}
\hline
\multicolumn{3}{c}{\textbf{DeepGlobe \cite{demir2018deepglobe} test-set}}                                  \\ \hhline{===}
\multicolumn{3}{c}{\textbf{Base Network CoANET mIoU (\%):  67.9 \%}}             \\ \hhline{===}
\multicolumn{1}{c|}{\textbf{Configuration}} & \multicolumn{1}{c|}{mIoU(\%)} & memory usage (GB) \\ \hline
\multicolumn{3}{c}{\textbf{Levels of cascading for Global step}}                 \\ \hline
\multicolumn{1}{c|}{(+) CascadePSP (1-level)} & \multicolumn{1}{c|}{$68.1_{\uparrow_{0.2}}$} & 1.03 \\ 
\multicolumn{1}{c|}{(+) CascadePSP (3-level)} & \multicolumn{1}{c|}{$68.8{\uparrow_{0.9}}$} & 1.03 \\ 
\multicolumn{1}{c|}{(+) PRN}                  & \multicolumn{1}{c|}{$\mathbf{70.6{\uparrow_{2.7}}}$} & \textbf{1.03} \\ \hline
\multicolumn{3}{c}{\textbf{Addition of Local step for different image crop sizes $L$}}              \\ \hline
\multicolumn{1}{c|}{(+) CascadePSP (L=512)}   & \multicolumn{1}{c|}{$68.9{\uparrow_{1.0}}$} & 2.12 \\ 
\multicolumn{1}{c|}{(+) CascadePSP (L=900)}   & \multicolumn{1}{c|}{$69.2{\uparrow_{1.3}}$} & 3.46 \\ 
\multicolumn{1}{c|}{(+) CascadePSP (L=1024)}  & \multicolumn{1}{c|}{$69.7{\uparrow_{1.8}}$} & 4.08 \\ 
\multicolumn{1}{c|}{(+) PRN}              & \multicolumn{1}{c|}{$\mathbf{70.6{\uparrow_{2.7}}}$} & \textbf{1.03} \\ \hline
\end{tabular}
}
\caption{\textbf{Quantitative results comparing \ours $(P=64)$ with CascadePSP \cite{cheng2020cascadepsp} on DeepGlobe test dataset}.}
  \label{tab:cascade}
\end{table} 
As an ablation experiment, we first consider just the global step of CascadePSP and change the number of cascade levels. This provides very
 marginal improvement over the base network and it is clearly outperformed by \ours (top half of Table \ref{tab:cascade}). Then we add the local step for CascadePSP and vary the image crop size parameter that it uses. This continues to improve the performance of CascadePSP, but it is still dominated by \ours (bottom half of Table \ref{tab:cascade}). The memory usage of CascadePSP grows with crop size and even when it needs 4 times as much memory as \ours, it is still outperformed by \ours.


\subsubsection{Saliency Detection on DUTS }
We next consider saliency detection (identifing the pixels of the salient objects in an image) using the DUTS dataset \cite{wang2017learning} and one of its state-of-the-art methods, 
PFAN \cite{zhao2019pyramid}, as the base network. The results are shown in Table \ref{tab:duts}. Adding \ours resulted in significant improvement of +3.8\% and +7.4\% in mIoU and mBA, again showing the potential of \ours in improving different kinds of networks. 

\begin{table}[htpb]
\centering
\begin{tabular}{lccc}
\hline
\multicolumn{4}{c}{\textbf{DUTS\cite{wang2017learning} test-set}} \\ \hhline{====}
\multicolumn{1}{l|}{\textbf{Methods}} &
  \multicolumn{1}{l|}{mIoU (\%)} &
  \multicolumn{1}{l|}{mBA (\%)} &
  \multicolumn{1}{l}{MAE} \\ \hline
\multicolumn{1}{l|}{\begin{tabular}[c]{@{}l@{}}RFCN \cite{wang2016saliency}\\ (+) \ours\end{tabular}} &
  \multicolumn{1}{c|}{\begin{tabular}[c]{@{}c@{}}52.8\\ $\mathbf{57.1_{\uparrow 4.3}}$\end{tabular}} &
  \multicolumn{1}{c|}{\begin{tabular}[c]{@{}c@{}}40.7\\ $\mathbf{48.5_{\uparrow 7.8}}$\end{tabular}} &
  \begin{tabular}[c]{@{}c@{}}0.0897\\ \textbf{0.0807}\end{tabular} \\ \hline
\multicolumn{1}{l|}{\begin{tabular}[c]{@{}l@{}}PFAN \cite{zhao2019pyramid}\\ (+) \ours\end{tabular}} &
  \multicolumn{1}{c|}{\begin{tabular}[c]{@{}c@{}}66.1\\ $\mathbf{69.9_{\uparrow 3.8}}$\end{tabular}} &
  \multicolumn{1}{c|}{\begin{tabular}[c]{@{}c@{}}51.2\\ $\mathbf{58.6_{\uparrow 7.4}}$\end{tabular}} &
  \begin{tabular}[c]{@{}c@{}}0.0452\\ \textbf{0.0386}\end{tabular} \\ \hline
\end{tabular}
\caption{\textbf{ \ours and  DUTS Saliency detection test dataset}.}
\label{tab:duts}
\end{table}


\subsubsection{Few-shot segmentation on FSS-1000 \cite{li2020fss}.}
Finally, in Table \ref{tab:fss}, we apply \ours to few-shot segmentation over the FSS-1000 \cite{li2020fss} dataset with EfficientLab \cite{hendryx2019meta} and ARN \cite{li2020fss} as the base networks. Again there is consistent improvement of at least
+1.3\% and +2.1\% in mIoU and mBA.
\begin{table}[htpb]
\centering
\resizebox{\columnwidth}{!}{%
\begin{tabular}{lcc}
\hline
\multicolumn{3}{c}{\textbf{FSS-1000 \cite{li2020fss} test-set}}                                    \\ \hhline{===}
\multicolumn{1}{c|}{\textbf{Methods}} & \multicolumn{1}{c|}{mIoU (\%)} & mBA (\%) \\ \hline
\multicolumn{1}{l|}{\begin{tabular}[c]{@{}l@{}}Adapted Relation Network \cite{li2020fss}\\ (+) PRN\end{tabular}} &
  \multicolumn{1}{c|}{\begin{tabular}[c]{@{}c@{}}80.1\\ $\mathbf{82.7_{\uparrow_{2.6\%}}}$\end{tabular}} &
  \begin{tabular}[c]{@{}c@{}}69.8\\ $\mathbf{72.9_{\uparrow_{3.1\%}}}$\end{tabular} \\ \hline
\multicolumn{1}{l|}{\begin{tabular}[c]{@{}l@{}}EfficientLab \cite{hendryx2019meta}\\ (+) PRN\end{tabular}} &
  \multicolumn{1}{c|}{\begin{tabular}[c]{@{}c@{}}82.8\\ $\mathbf{84.1_{\uparrow_{1.3\%}}}$\end{tabular}} &
  \begin{tabular}[c]{@{}c@{}}71.1\\ $\mathbf{73.2_{\uparrow_{2.1\%}}}$\end{tabular} \\ \hline
\end{tabular}
}
\caption{\textbf{\ours and FSS-1000 Saliency detection test dataset}.}
\label{tab:fss}
\end{table}

\section{More qualitative results for performance evaluation of PRN}
\begin{figure*}[htpb]
    \centering
    \includegraphics[width=\linewidth]{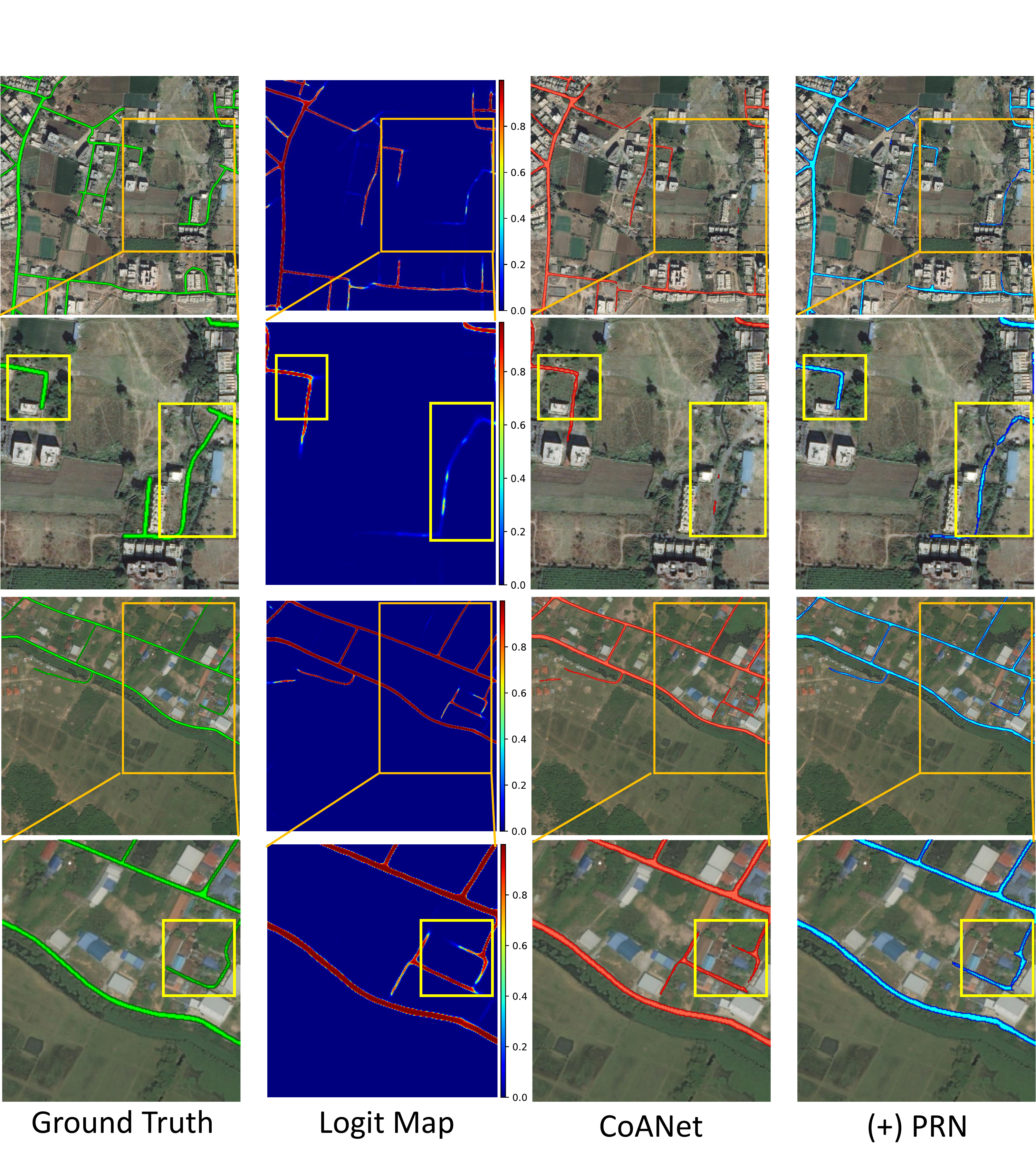}
    \caption{\textbf{Qualitative results produced by \ours $(P=64)$ on samples from DeepGlobe test dataset.} Left to Right: Ground Truth, Logit map from base network CoANet \cite{21TIP-Road}, Prediction from CoANet , Refined by \ours. Yellow boxes denote regions of refinement.}
    \label{fig:supp_deepglobe}
\end{figure*}

\begin{figure*}[htpb]
    \centering
    \includegraphics[width=0.92\linewidth]{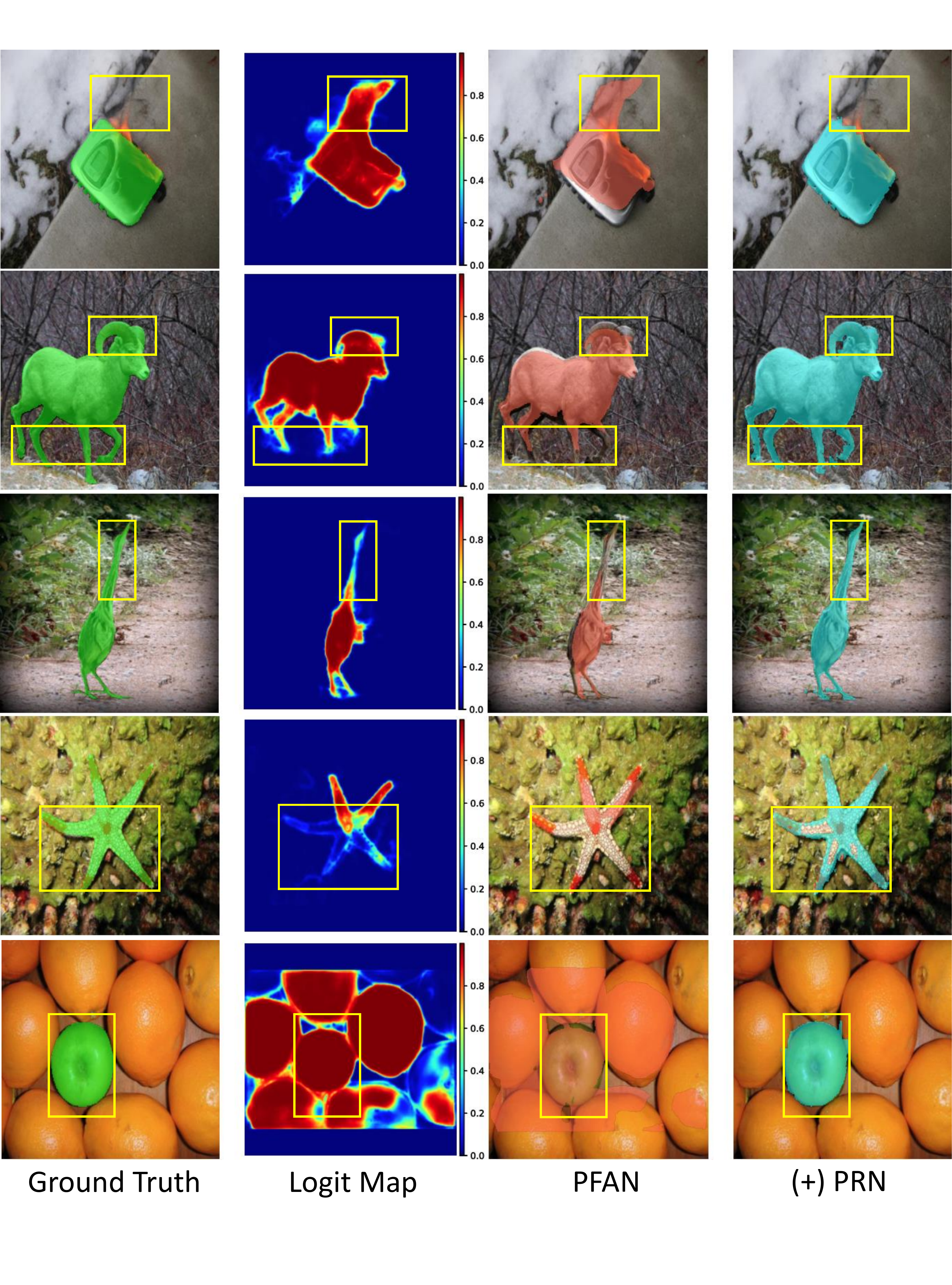}
    \caption{\textbf{Qualitative results produced by \ours $(P=64)$ on samples from DUTS saliency detection test dataset.} Left to Right: Ground Truth, Logit map from base network PFAN \cite{zhao2017pyramid}, Prediction from PFAN, Refined by \ours. Yellow boxes denote regions of refinement.}
    \label{fig:supp_duts}
\end{figure*}

\begin{figure*}[htpb]
    \centering
    \includegraphics[width=\linewidth]{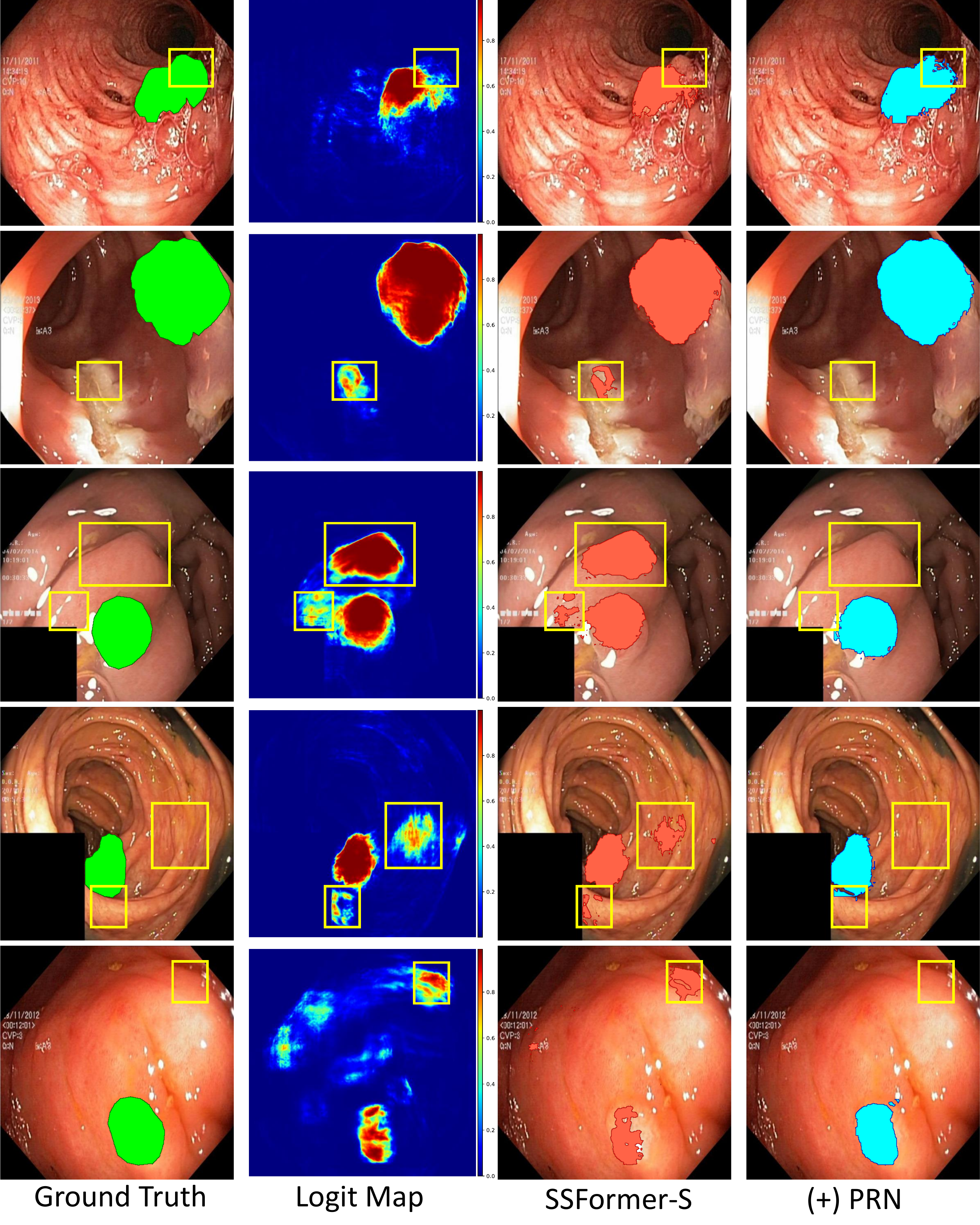}
    \caption{\textbf{Qualitative results produced by \ours $(P=64)$ on samples from Kvasir-SEG test dataset.} Left to Right: Ground Truth, Logit map from base network SSFormer-S \cite{wang2022stepwise},  Prediction from SSFormer-S, Refined by \ours. Yellow boxes denote regions of refinement.}
    \label{fig:supp_kvasir}
\end{figure*}

Figures \ref{fig:supp_deepglobe}, \ref{fig:supp_duts} and \ref{fig:supp_kvasir} show qualitative results for PRN refinement \ours with a patch-size parameter $P=64$ when applied to DeepGlobe, DUTS saliency detection and Kvasir-SEG datasets respectively. Going from left to right, the figures show Ground Truth, Logit map from the base network, binarized prediction of the base network, and PRN refined map.
Yellow boxes mark the regions of refinement. It can be observed that \ours is able to achieve significant visual improvement compared to the corresponding segmentation maps from state-of-the-art base networks with fixed (0.5) thresholding.

\begin{figure*}[htpb]
    \centering
    \includegraphics[width=\linewidth]{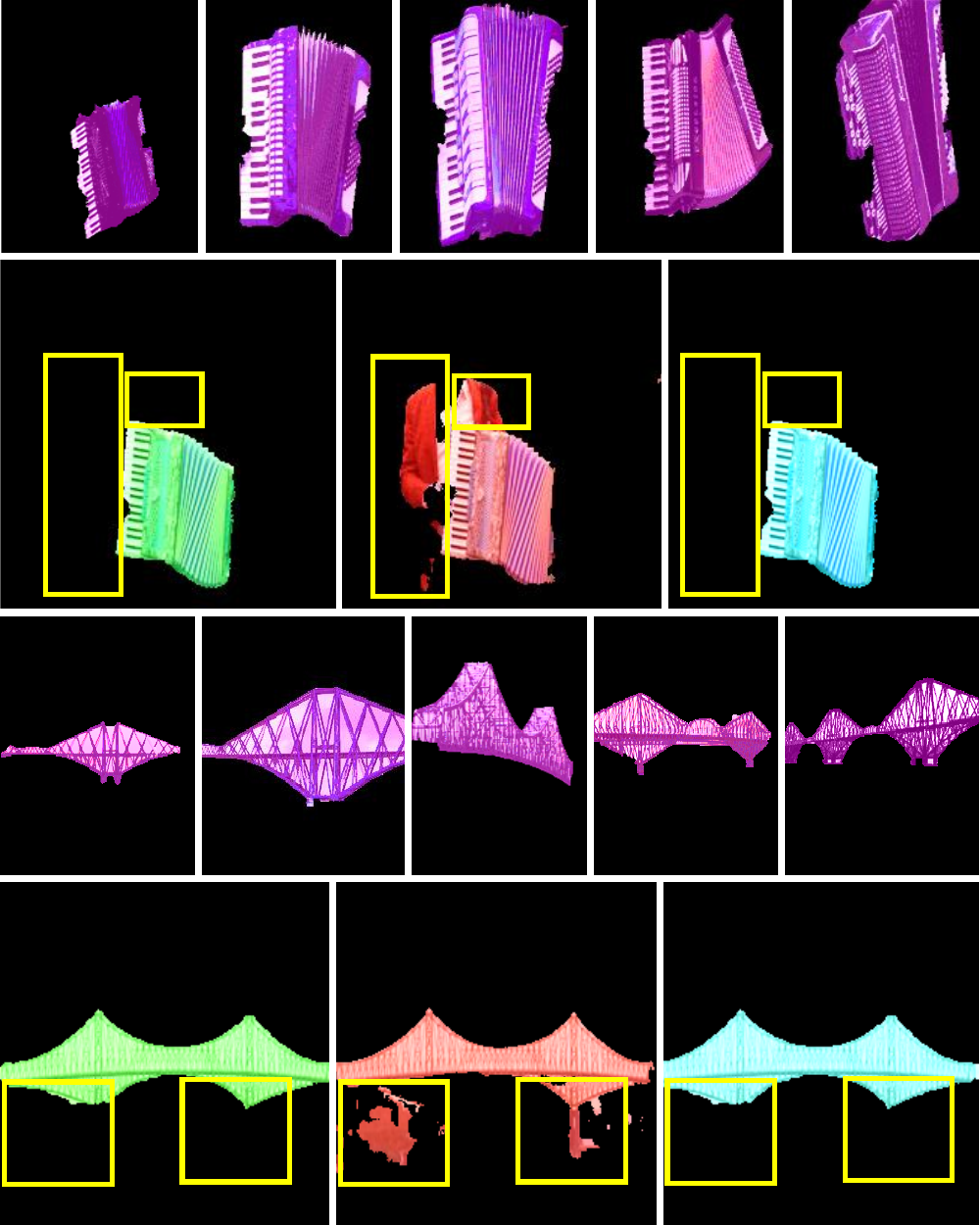}
    \caption{\textbf{Qualitative results produced by \ours ($P=64$) on samples from FSS-1000 test dataset}. Rows 1\& 3: Support images. Left to right: Ground Truth, Thresholded (0.5) prediction from Adapted Relation Network \cite{zhao2019pyramid}, Refined map by PRN. Yellow boxes denote regions of refinement.}
    \label{fig:qual_fss}
\end{figure*}
\section{Conclusion}
We proposed PatchRefineNet (\ours), a post-processing network that sits on top of a base segmentation model and learns to correct its spatial biases. \ours uses a novel learning signal that is computed from binarizing each patch separately.  \ours complements virtually any binary segmentation network and also works with saliency detection. In our experiments across different base models, \ours consistently helps the base networks improve both mIoU 
and mBA by over 2-3 \%. \footnote{Additional experiments can be found in the supplemenatry material.}

{\small
\bibliographystyle{ieee_fullname}
\bibliography{egbib}
}

\end{document}